\documentclass{article}

\usepackage{microtype}
\usepackage{graphicx}
\usepackage{booktabs} % for professional tables

\usepackage{float}
\usepackage{graphicx}
\usepackage{subcaption}
\usepackage{booktabs} % For better table lines
\usepackage{multirow} % For multirow cells

\usepackage{hyperref}

\usepackage[accepted]{icml2025}

\usepackage{amsmath}
\usepackage{amssymb}
\usepackage{mathtools}
\usepackage{amsthm}
\usepackage{dblfloatfix}
\usepackage[capitalize,noabbrev]{cleveref}

\theoremstyle{plain}

\theoremstyle{definition}

\theoremstyle{remark}

\usepackage[textsize=tiny]{todonotes}
\usepackage{graphicx}

\icmltitlerunning{CuPID: Leveraging Masked Single-Lead ECG Modelling for Enhancing the Representations}
\usepackage{acronym}
\usepackage[acronym]{glossaries}
\newacro{SSL}{Self-Supervised Learning} 
\newacro{CHRONOS}{Contrasting Heads Represent Opposed Natures of Signals}
\newacro{ML}{Machine Learning}
\newacro{ECG}{Electrocardiogram} 
\newacro{AFib}{Atrial Fibrillation}
\newacro{DEBS}{Distilled Encoding Beyond Similarities}
\newacro{EEG}{Electroencephalogram}
\newacro{CPC}{Contrastive Predictive Coding}
\newacro{PCLR}{Patient Contrastive Learning}
\newacro{EMA}{exponential moving average}
\newacro{SBnCL}{Subject-Based non Contrastive Learning}
\newacro{SVC}{ Support Vector Classificatier}
\newacro{BYOL}{Boostrap Your Own Latent}
\newacro{DINO}{Self-Distillation with no Labels}
\newacro{PAR}{Pondered Average Representation}
\newacro{SHHS}{Sleep Heart Health Study}
\newacro{ViT}{Vision Transformer}
\newacro{MLP}{Multilayer Perceptron}
\newacro{MIT-ARR}{MIT-BIH Arrhythmia Database}
\newacro{MIT-AFIB}{MIT-BIH Atrial Fibrillation}
\newacro{PCA}{Principal Component Analysis}
\newacro{MAE}{Masked Autoencoders}
\newacro{BYOL}{Bootstrap Your Own Latent}
\newacro{ReLU}{rectified linear unit}
\newacro{PSG}{Polysomnography}
\newacro{SOTA}{state-of-the-art}
\newacro{TF-C}{Time-Frequency Consistency}
\newacro{Cinc2017}{Physionet Challenge 2017}
\newacro{NSRR}{National Sleep Research Resource}
\newacro{CLOCS}{Contrastive Learning of Cardiac Signals Across Space}
\newacro{DEAPS}{Distilled Embedding for Almost-Periodic Time Series}
\newacro{SOTA}{state-of-the-art}
\newacro{VIC-REG}{Variance-Invariance-Covariance Regularization}
\newacro{SGD}{Stochastic Gradient Descent}
\newacro{LOO}{Leave-One-Out}
\newacro{DIVA}{Disentangling Invariant and tempo-Variant Attributes}
\newacro{SIE}{Split Invariant-Equivariant}
\newacro{MIT-PSG}{MIT-BIH Polysomnographic}
\newacro{SHAP}{SHapley Additive exPlanations}
\newacro{GRU}{Gated Recurrent Unit}
\newacro{PLITA}{Parallel-Learning of Invariant and Tempo-variant Attributes}
\newacro{ISL}{Intra-inter Subject Self-Supervised Learning}
\newacro{ST-MEM}{Spatio-Temporal Masked Electrocardiogram Modeling}
\newacro{ASTCL}{Adversarial Spatiotemporal Contrastive Learning}
\newacro{CuPID}{Cueing the Predictor Increments the Detailing}
\newacro{FFT}{Fast Fourier Transform}
\newacro{MDM}{Masked Data Modelling}
\newacro{MSE}{Mean Squared Error}
\newacro{EBM}{Energy-Based Modelling}
\newacro{LT-AF}{Long Term AF}
\newacro{MIT-SVA}{MIT-BIH Supraventricular Arrhythmia}
\newacro{NSR}{Normal Sinus Rhythm}
\newacro{SiamMAE}{Siamese Masked Autoencoders}
\newacro{MAE}{Masked Autoencoders}
\newacro{BERT}{Bidirectional Encoder Representations from Transformers}
\newacro{I-JEPA}{Image-based Joint-Embedding Predictive Architecture}
\newacro{NLP}{Natural Language Processing}
\newacro{MTAE}{Masked Time Autoencoder}
\newacro{MLAE}{Masked Lead AutoEncoder}
\newacro{MaeFE}{MAE family of ECG}
\newacro{MLTAE}{Masked Lead and Time Autoencoder}
\newacro{CMSC}{Contrastive Multi-segment Coding}
\newacro{HRV}{Heart Rate Variability}
\newacro{ST-MEM}{Spatio-Temporal Masked Electrocardiogram Modeling}

\begin{document}

\twocolumn[
\icmltitle{CuPID: Leveraging Masked Single-Lead ECG Modelling \\ for Enhancing the Representations}



\icmlsetsymbol{equal}{*}

\begin{icmlauthorlist}
\icmlauthor{Adrian Atienza}{yyy}
\icmlauthor{Gouthamaan Manimaran}{yyy}
\icmlauthor{Jakob E. Bardram}{yyy}
\icmlauthor{Sadasivan Puthusserypady}{yyy}
\end{icmlauthorlist}

\icmlaffiliation{yyy}{Department of XXX, University of YYY, Location, Country}

\icmlcorrespondingauthor{Firstname1 Lastname1}{first1.last1@xxx.edu}
\icmlcorrespondingauthor{Firstname2 Lastname2}{first2.last2@www.uk}

\icmlkeywords{Machine Learning, ICML}

\vskip 0.3in
]




\begin{abstract}
Wearable sensing devices, such as \ac{ECG} heart-rate monitors, will play a crucial role in the future of digital health.
This continuous monitoring leads to massive unlabeled data, incentivizing the development of unsupervised learning frameworks.
%
%associate these single-lead \ac{ECG} signals to their anticipated clto inical outcomes.
%
While \ac{MDM} techniques have enjoyed wide use,
their direct application to single-lead ECG data is suboptimal due to the decoder's difficulty handling irregular heartbeat intervals when no contextual information is provided.
In this paper, we present \acf{CuPID}, a novel \ac{MDM} method tailored to single-lead ECGs.
\ac{CuPID} enhances existing \ac{MDM} techniques by cueing spectrogram-derived context to the decoder, thus incentivizing the encoder to produce more detailed representations. 
This has a significant impact on the encoder's performance across a wide range of different configurations, leading
\ac{CuPID} to outperform state-of-the-art methods in a variety of downstream tasks. 

\end{abstract}

\section{Introduction}

The wearable sensing field has seen remarkable advancements in recent years, 
% By enabling real-time data collection on physiological parameters, these devices will play a 
and is expected to play a crucial role in the future of digital health. 
One widely used type of wearable health sensor is the heart monitor that captures cardiac activity as single-lead \acf{ECG} signals during free-living conditions, such as in the patient's home.
Mapping these signals with significant clinical outcomes has the potential to provide outstanding benefits such as simplifying the diagnostic process \citep{s_ecg_1} or enabling users to engage proactively in tracking their cardiac health \citep{s_ecg_2}.
%(i) Simplifying the diagnostic process and minimizing the necessity for comprehensive testing \citep{s_ecg_1}, and (ii) Enabling users to engage proactively in tracking their heart health by offering instant access to health data and insights \citep{s_ecg_2}.
%
In this context, models that extract information from single-lead \ac{ECG} into generalizable representations are mandated to address distinct downstream tasks. 
In addition, this continuous monitoring leads to massive unlabeled datasets. This makes \acf{SSL} framework particularly well-suited for addressing this clinical challenge.

\begin{figure}[t]
\centering
\fbox{\includegraphics[width=0.8\linewidth]{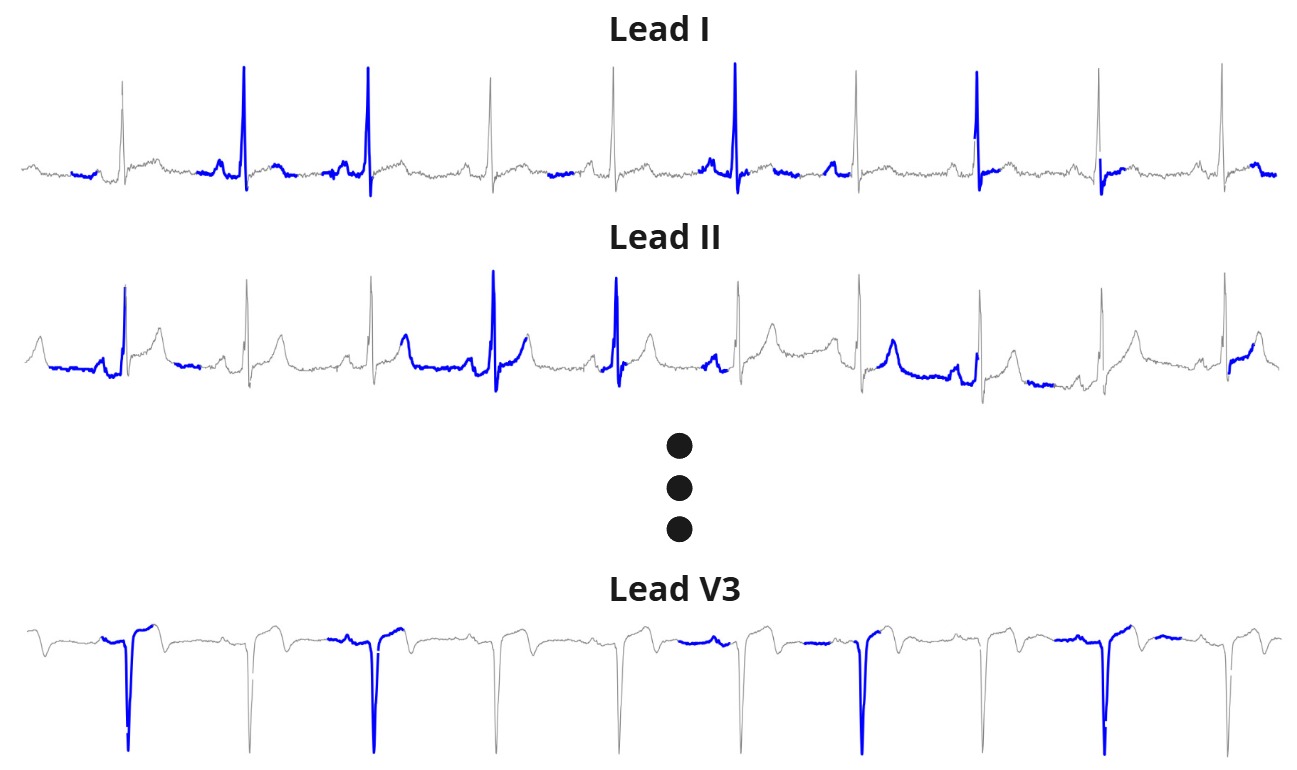}}
\caption{Example of the commonly used masking strategy proposed in \cite{stmem} for 12-Lead ECG processing. Unmasked portions are displayed in blue. Most of the ECG waves remain unmasked in at least one of the leads.}
\label{fig:stmem-masking}
\end{figure}

Recently, \acf{MDM} methods have been gaining attention in the \ac{SSL} field \citep{mae, siam_auto, jepa}.
They rely on masking a portion of the input and driving a transformer-based encoder, typically a \ac{ViT} \citep{vit} to compute detailed representations that enable a decoder to infer the unseen patches. 
In the realm of 12-lead \ac{ECG} processing, \ac{MDM} methods have been applied recently with promising results \cite{stmem}.
They prove that independently masking the different leads outperforms the strategy of consistently masking across time proposed in \ac{MTAE} \cite{maefe}.

\begin{figure*}[t]
    \centering
    \subfloat[]{\fbox{\includegraphics[width=0.46\textwidth]{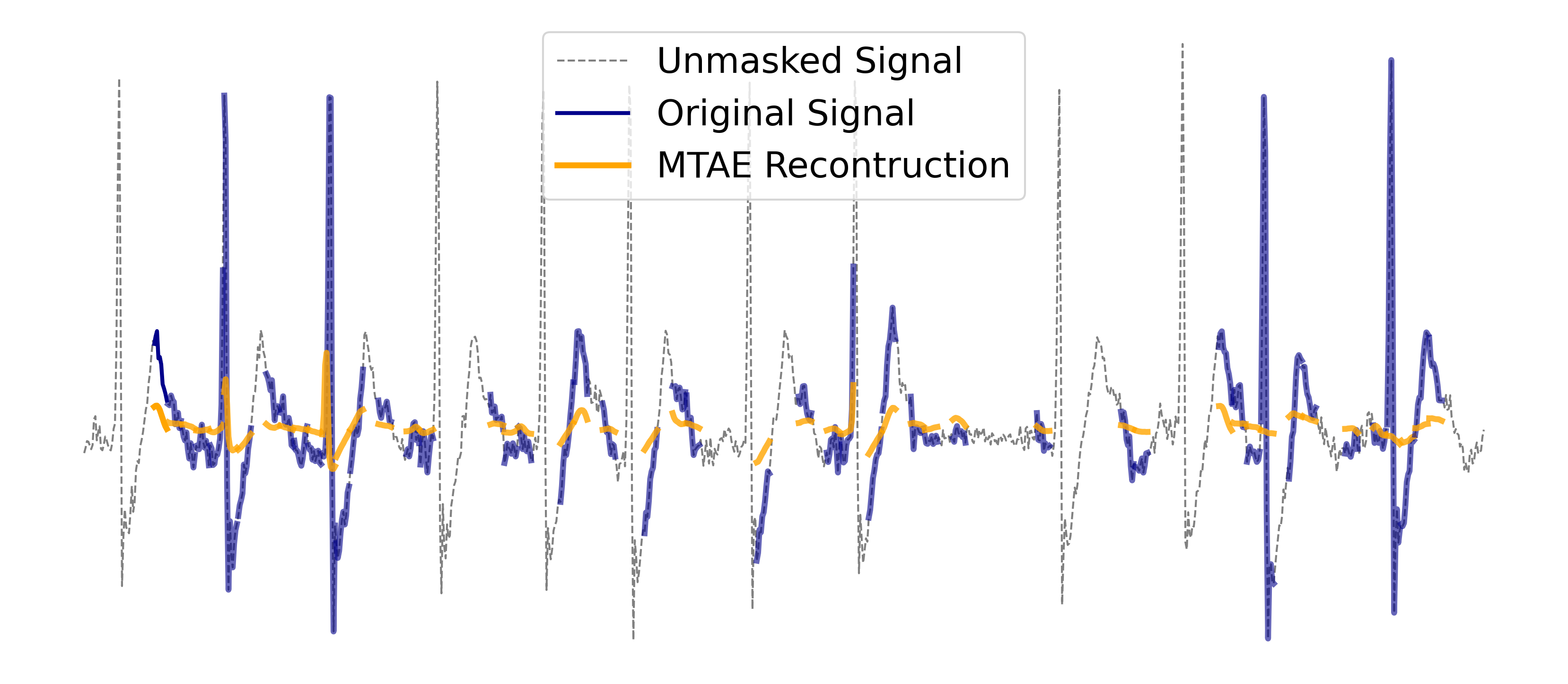}}\label{fig:baseline_rec}}
    \hfill
    \subfloat[]{\fbox{\includegraphics[width=0.48\textwidth]{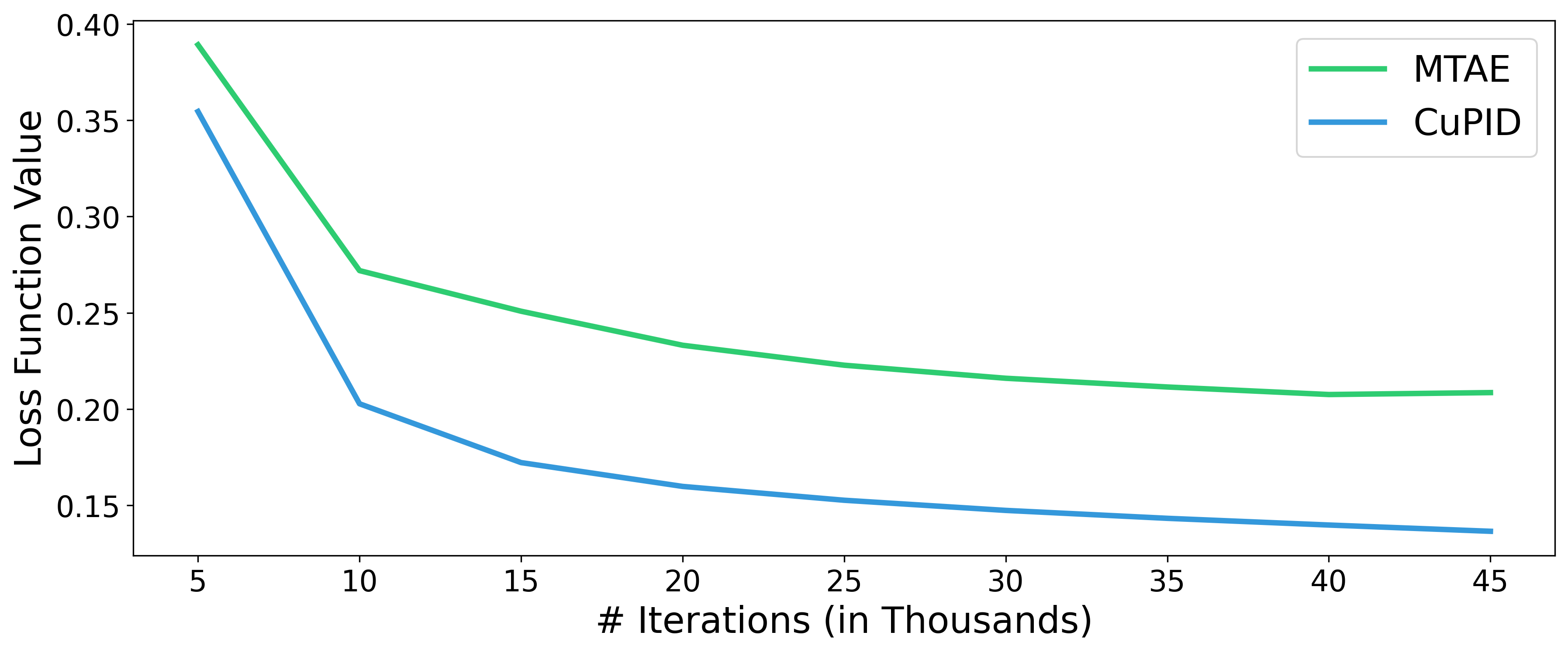}}\label{fig:loss}}
    \\
     \subfloat[]{\fbox{\includegraphics[width=0.46\textwidth]{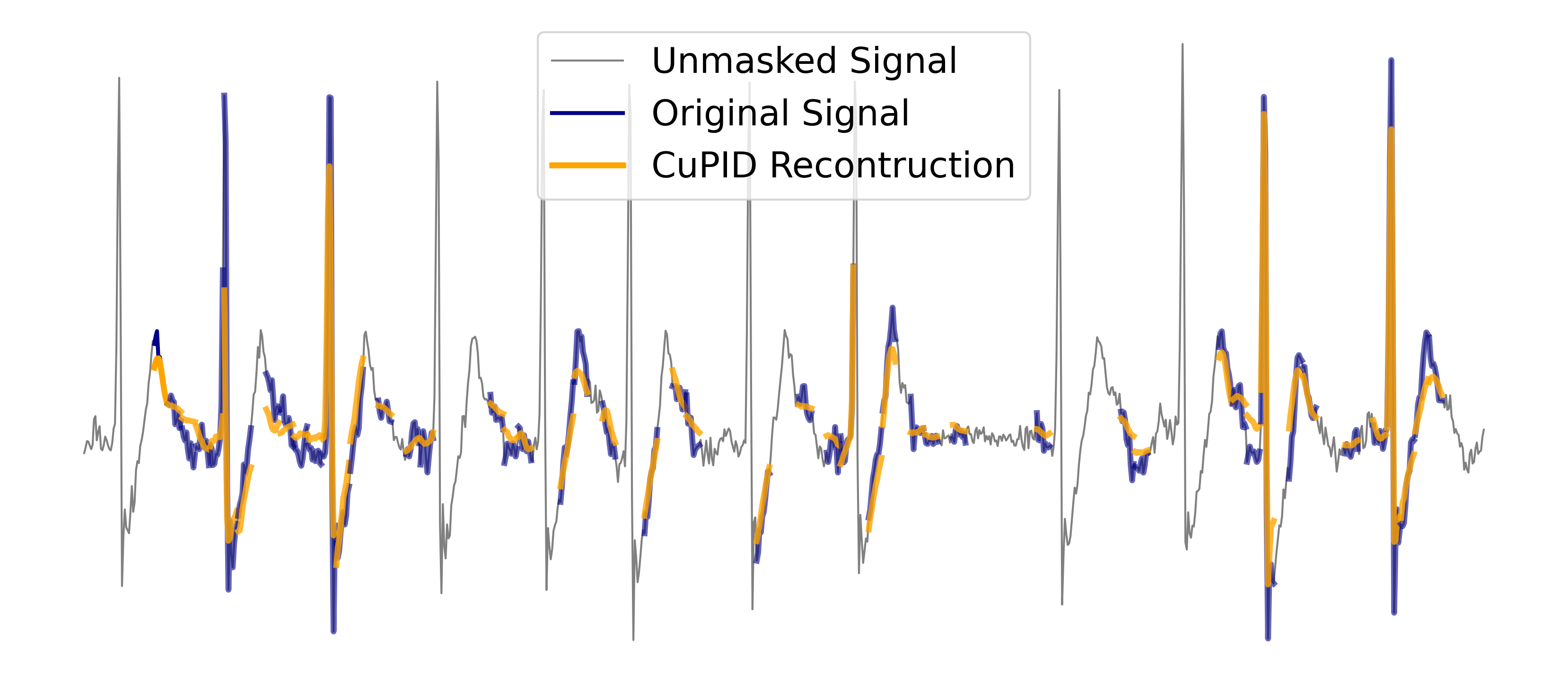}}\label{fig:cupid_rec}}
    \hfill
    \subfloat[]{\fbox{\includegraphics[width=0.48\textwidth]{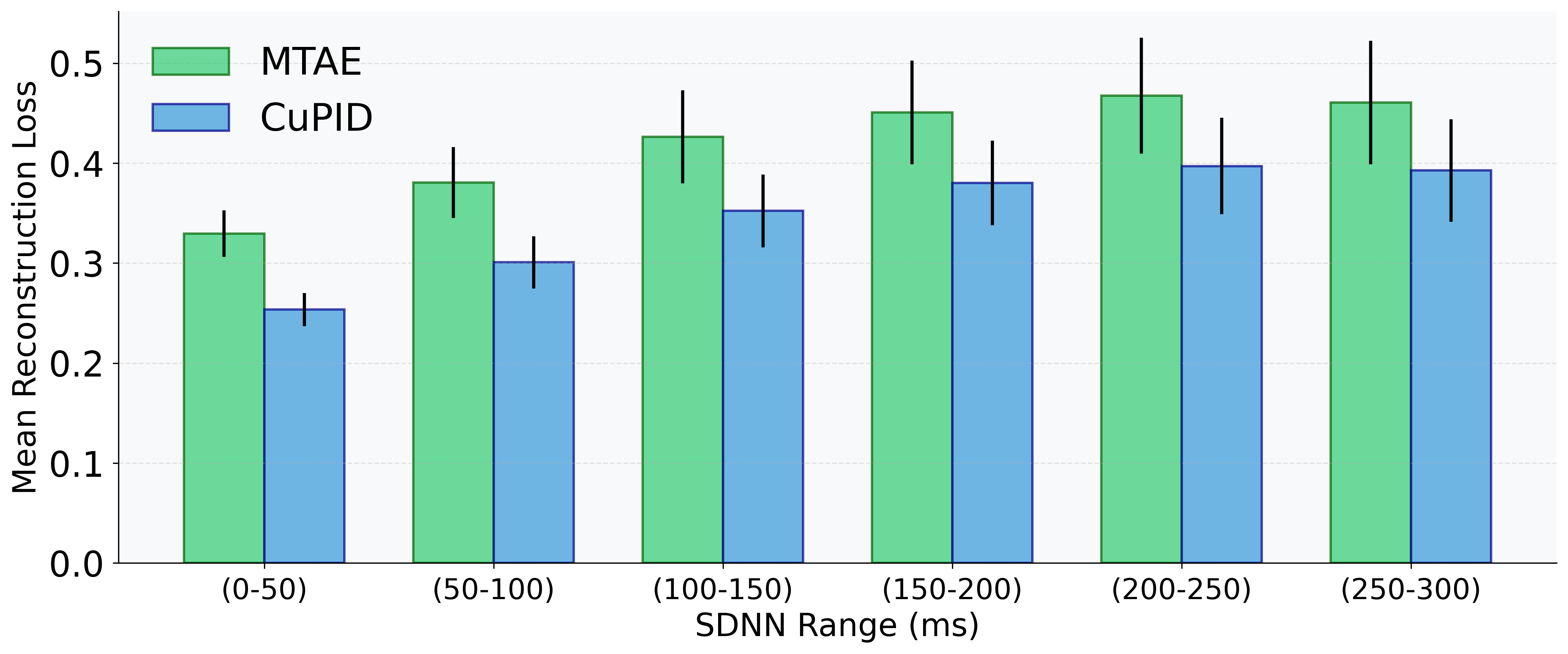}} \label{fig:hrv}} 
    \caption{
    Reconstruction comparison between CuPID and MTAE.
    Figures \ref{fig:baseline_rec} and \ref{fig:cupid_rec} display a reconstruction from MTAE and \ac{CuPID}, where the unmasked part, the ground truth, and the inference computed by both methods are represented in gray, blue, and orange, respectively.
    %
    %Figure \ref{fig:loss} illustrates that CuPID achieves significantly lower values for the loss function.
    Figure \ref{fig:loss} illustrates loss function evolution across pre-training.
    Figure \ref{fig:hrv} compares the performance of CuPID and MTAE, across different heartbeat irregularity levels, measured in SDNN (Standard Deviation of Normal Intervals).
    }
    \label{fig:2x2grid}
\end{figure*}

% One can think that preserving the time mask over the leads, as proposed in \ac{MTAE} \cite{maefe} would make the task more challenging, driving the encoder to learn better the data. 
% In theory, reconstructing the missing portions of an ECG signal might appear straightforward because of its periodic nature.
%
% Independently masking the different leads, as shown in Figure \ref{fig:stmem-masking}, 
This lead-independent masking strategy, represented in Figure \ref{fig:stmem-masking}, allows the encoder to model the temporal positions of the distinct ECG waves by integrating the unmasked portions of the various leads. 
This contextual information is crucial, as it enables the decoder to deal with fluctuations in the time intervals between heartbeats.
However, this masking strategy can not be applied to single-lead ECG data, our data of interest, which just accommodates one signal.
The decoder encounters challenges in handling irregular heartbeat intervals, as it lacks contextual information from other leads.
Since not inferring exactly this position has a big impact on the loss, the decoder is cautious when reconstructing the masked patches. 
Figure \ref{fig:baseline_rec} displays how it estimates a value near the average rather than matching precisely the signal’s morphology.

    % \item These high HRV strips are the ones that are truly interesting to be modeled, since bla bla bla.

This paper presents \acf{CuPID}, which is a novel \ac{SSL} method that addresses the issue mentioned above by cueing the decoder with contextual information provided by the spectrogram of the input signal. 
The spectrogram is expected to mirror the contextual information provided by other leads in the 12-Lead ECG framework.
It is fed into the attention mechanism of the decoder as the Key (K) to ensure that its role is merely informative and its value can not be used directly to reconstruct the representations.
It leads to a significant decrease in the loss function values, as shown in Figure \ref{fig:loss},
and reconstructions are adjusted more to the morphology of the original signal, as captured in Figure \ref{fig:cupid_rec}.
Figure \ref{fig:hrv} shows how this behavior is consistent across different levels of irregularity in heartbeat intervals.\\

%
%So far, we only have demonstrated that \ac{CuPID} accomplishes the learning objective more successfully. 
%
Although these results are insignificant on their own since \ac{CuPID}'s decoder is provided with additional information, we hypothesize:
(I) The decoder’s inability to reconstruct the original signal due to the irregularity in the heartbeat intervals limits the encoder’s learning potential.
(ii) Cueing the decoder with the spectrogram endows it with the potential to solve the task. It drives the encoder to compute detailed patch representations which can be used to accurately reconstruct the input.
(iii) The more informative the patch representations are, the more performant the encode will be at the time of addressing downstream tasks.\\
% Evaluation
To assess our hypothesis, we have conducted an extensive evaluation where \ac{CuPID} is compared against the existing state-of-the-art \ac{SSL} methods tailored for single-lead ECG analysis. 
%
%\cite{clocs, pclr, mix_up} in three relevant downstream. 
%
Up to three distinct databases; \ac{MIT-AFIB} \citep{mit-afib}, Physionet Challenge 2017 \cite{physio_challenge}, and \ac{LT-AF} \citep{l-af-dataset}, are considered.  
Remarkably, \ac{CuPID} achieves significantly superior performance when compared with single-lead ECG methods.
Finally, the benefit of incorporating the spectrogram compared with the MTAE baseline is assessed for different configurations.\\

In summary, the contributions of this paper are: 
\begin{itemize}
\item We have discussed the limitations of applying \ac{MDM} techniques directly to single-lead \ac{ECG} signals due to the idiosyncrasy of this kind of data.

\item We introduce \ac{CuPID}, a novel \ac{SSL} method that addresses these limitations by helping the decoder during the pre-training. This is made by incorporating the spectrogram of the input signal to the attention mechanism as the Key, limiting its role to be merely informative. 

\item  We provide a model that achieves markedly enhanced results in a variety of downstream tasks that are relevant for cardiovascular remote monitoring. 

\end{itemize}

\section{Related Work}
\subsection{\acf{MDM}}
\acf{MDM} has been a commonly used technique in the \ac{NLP} field. Methods such as \ac{BERT} \citep{bert} that rely on hiding a series of words within a sentence and optimizing a decoder to infer these words have proven to be the most effective pre-training method in the field. 

In recent times, this pre-training mechanism has been adapted in the field of computer vision.
Existing methods, such as, \ac{MAE} \citep{mae} or \ac{SiamMAE} \citep{siam_auto} incorporate a decoder trained to reconstruct masked patches from the original input. This approach has shown promising results in the field of computer vision, outperforming gold-standard \ac{EBM} methods such as \ac{VIC-REG} \citep{vicreg}, \ac{DINO} \citep{dino}, or \ac{BYOL} \citep{byol}.

\subsection{SSL in 12-Lead \ac{ECG} Signal Processing}
In the realm of 12-lead ECG signals, research has effectively utilized the \ac{MDM} framework. 
The availability of various leads broadens the scope for the strategy of input masking.
Techniques like MTAE, MLAE, and MLTAE, all introduced by \ac{MaeFE} \citep{maefe}, suggest three masking strategies: temporal masking, spatial masking across different leads, or a combination of both. \ac{ST-MEM} \citep{stmem} reaches the state-of-the-art performance by employing a joint decoder that reconstructs the original input attending to each lead independently. 
It proves that adopting this combined strategy by independently masking the different leads achieves the best results.

Among the four listed methods, only MTAE is suitable for single-lead ECG signals, as the other three require multiple leads. MTAE is not only included in the evaluation, but also in the ablation studies since this method is CuPID's analogous version without the use of the spectrogram.

\subsection{SSL in Single-Lead \ac{ECG} Signal Processing}%
Most-widely used single-lead \ac{ECG} \ac{SSL} methods follows a \ac{EBM} approach;  
(i) \ac{CLOCS} \citep{clocs} utilizes two consecutive \ac{ECG} time strips as positive pairs, 
(ii) Mixing-Up~\citep{mix_up} introduces a more tailored data augmentation product of two time series from the same recording,
(iii) \ac{PCLR} \citep{pclr} which considers two time strips from the same subject but different recordings.
While all these methods utilize the Contrastive Learning \citep{simclr} as a common framework for learning the invariant attributes considering non-overlapping inputs as positive pairs, \ac{DEAPS} \citep{deaps} follows a non-contrastive learning approach.
(iv) It drives the model to capture the also dynamic patterns of the single-lead ECGs.  

All of these \ac{SSL} methods will compose the set of baselines for the \ac{CuPID}'s evaluation, where the representations computed by each pre-trained model will be employed for addressing several downstream tasks.

\subsection{Use of Spectogram for ECG Processing}
Spectrograms have enjoyed wide use in ECG signal processing due to their ability to provide a time-varying spectral density description of the data. Consequently, it is often used as a substitute for the ECG signal as input for deep learning models designed to address various arrhythmia classification tasks \cite{spec1, spec2}. Additionally, the spectrogram can be treated as an image, enabling the application of SSL techniques developed for computer vision to this type of data \cite{spec_ssl}.

All the previously mentioned methods actively utilize the spectrogram to map input to clinical outcomes. CuPID stands out since the spectrogram is only employed during pre-training to provide the decoder with the temporal context of the \ac{ECG} waves. The encoder does not receive the spectrogram, and thus, it is not used during inference.

\section{\acf{CuPID}}

\begin{figure*}[ht]
\centering
\fbox{\includegraphics[width=\textwidth]{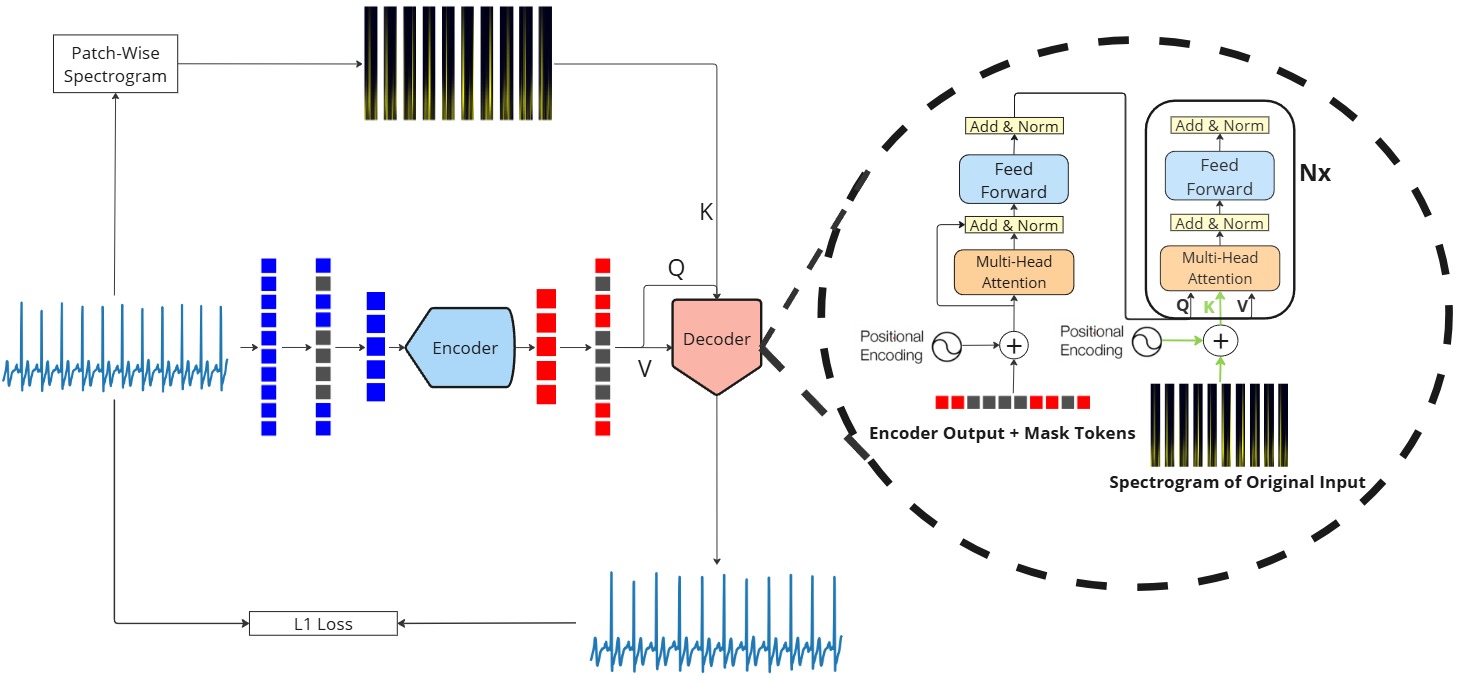}}
\caption{CuPID architecture. 
The left side of the Figure shows how the spectrogram is incorporated into the decoder's attention mechanism. This incorporation sets \ac{CuPID} apart from the standard framework for \ac{MDM}. The encoder is the model used to address the downstream tasks, while the decoder is discarded after the pre-training. Therefore, this spectrogram is not provided during the evaluation. 
The right side of the diagram provides a closer look at CuPID's decoder. Due to the challenges of using the spectrogram as a Key, the spectrogram is incorporated from the second block of the decoder. Its first block mirrors the standard decoder block for \ac{MDM} framework.}
\label{fig:cupid}
\end{figure*}

The core idea behind \ac{CuPID} is cueing the decoder with the contextual information provided by the spectrogram. Its workflow is illustrated in Figure \ref{fig:cupid}.
From left to right the original signal input is patched and embedded using a linear layer. 
A portion of these tokens (Represented as gray blocks in the figure) is randomly masked with a fixed ratio.
Only the unmasked tokens are passed through the encoder.
Learnable mask tokens with their respective positional encoding are placed in the original position of the masked segments.
What sets CuPID apart is that it uses the spectrogram as the Key for the attention mechanism, as represented in Figure \ref{fig:cupid}.
This decoder reconstructs the original input. The $\mathcal{L}_{1}$ metric is computed between this reconstruction and the original input. This loss function is only calculated on the masked patches.
It is important to note that the decoder is discarded after training, with the encoder being used for downstream tasks. Therefore, the spectrogram is only utilized during pre-training and not during inference.

\subsection{Role of Spectrogram for the decoder}
Due to the lack of contextual information, the baseline decoder faces challenges when handling irregular heartbeat intervals. 
The core idea behind CuPID is to provide the decoder with the needed contextual information from an alternative source. 
The spectrogram is identified as an effective tool for this purpose.
This section delves deeper into the rationale behind this decision and its implementation.

\paragraph{Spectrogram as Choice:}
A spectrogram is a visual representation of the spectrum of frequencies in a signal as they vary over time.
They are commonly generated using the \ac{FFT}, which converts a time-domain signal into its frequency components.

We identify the spectrogram as a tool that has the potential to provide the decoder with the needed contextual information since:
(i) The spectrogram retains the temporal aspect of the signal, allowing us to isolate the frequency component of each patch by dividing it into consistent patches aligned with the original input ones.
(ii) The distinct waves accommodated in the \ac{ECG} signal operate in distinct frequencies. 
This feature is leveraged by traditional signal processing methods to perform \ac{ECG} signal delineation \citep{wt_signal}.
Therefore, by providing the frequency component of each mask token, the decoder can determine if a wave is present there and identify the kind of wave it should reconstruct.

\paragraph{Limiting the Information Provided by the Spectrogram:}
Just as the time domain input is transformed into the frequency domain when computing the spectrogram, 
it can also be converted back to the time domain.
It means that the decoder could potentially reconstruct the original input without using the encoder’s representations. 
To prevent this, the spectrogram is used just as the K in the attention mechanism when fed into the decoder. 
This transformer-based decoder relies on the standard attention mechanism formulated on \cite{transformer}. It is composed of three components, i.e., query ($Q$), key ($K$), and value ($V$) and it is expressed as the following:

\begin{equation}
\operatorname{Attention}(Q, K, V)=\operatorname{softmax}\left(\frac{Q K^T}{\sqrt{d_k}}\right) V,
\end{equation}
where the query ($Q$) refers to the token that is attending the others for information,
the key ($K$) represents what information can be found in the specific token,
and the value ($V$) accommodates the information.
It is worth highlighting that the $K$ only has the potential of informing what kind of information it could be found in the respective token, but not providing information by itself. This information is provided by the corresponding $V$. 
In other words, even though the spectrogram gathers all the information needed for reconstructing the input, this information can not be applied directly.

\paragraph{Challenges of Using the Spectrogram as the Key:}
CuPID does not mask the spectrogram when fed into the decoder. 
The purpose of incorporating the spectrogram is to enhance mask tokens with contextual information about the temporal location and the kind of ECG waves each accommodates for an accurate reconstruction. Therefore, masking the spectrogram would be counterproductive.

However, a primary issue arises when using the spectrogram without masking it. 
Since this spectrogram is fed as the \textit{K} into the attention mechanism of the decoder, it cannot distinguish between informative tokens and mask tokens, as this distinction is not present in the spectrogram. 
Therefore, a token might be retrieving information from a mask token even when it is merely a mask and contains no actual information. 
To overcome this issue, \ac{CuPID} delays incorporating the spectrogram into the decoder’s second block. 
Consequently, the regular concatenation of encoder representations and mask tokens is used as the \textit{K} in the first encoder block. 
By doing this, \ac{CuPID} ensures that each mask token retains some information after the initial block, which can then be distilled in subsequent blocks with the context information provided by the spectrogram.

\paragraph{Decoder Workflow and Loss Function:}
Considering these two crucial aspects, Figure \ref{fig:cupid} depicts the \ac{CuPID} decoder. 
In the initial block, the inference follows a conventional approach, while the spectrogram is integrated into subsequent blocks as the $K$ in the attention mechanism. 
This decoder computes the single-lead \ac{ECG} reconstruction, which is compared to its corresponding original input using the $\mathcal{L}_{1}$ metric. 
This metric serves as the sole loss function of the model and is represented by the following formula:

\begin{equation}
\mathcal{L}_{1}(X, \hat{Y}, \mathcal{M}) = \frac{1} {sum(\mathcal{M})} \cdot
\sum_{i=1}^n\left|Y_i-\hat{Y}_i\right| \cdot \mathcal{M}_i,
\end{equation}
where $X$, $\hat{Y}$, $\mathcal{M}$, and $n$ represent the original input, the decoder reconstruction, the mask, and the number of patches.

\subsection{Implementation Details}
To ensure the replication of the method, we meticulously outline the hyperparameter configuration.\\

%To ensure the reproducibility of the method, this subsection details the configuration of the hyperparameters as well as the model architecture. 

\noindent \textbf{Model Architecture:} 
As is standard practice when evaluating MDM approaches, the standard ViT \cite{vit} architecture is used as the encoder module.
To align the encoder size with the data complexity and meet the efficiency requirements for constant monitoring, CuPID introduces a very lightweight model.
In detail, the \ac{ViT} architecture proposed by \ac{CuPID} for processing the single-lead \ac{ECG} signals count with four regular transformer blocks with four heads each and a model dimension of 128.
The input consists of a one-dimensional 10-second signal sampled at 100 Hz. 
This signal is split into patches with a length of 20 samples.
The influence of the patch size is studied in Section \ref{sec:abla}.
\\ 

\noindent \textbf{\ac{CuPID} Implementation and Optimization}: The decoder consists of a ViT model with two blocks and a dimension of 128. 
The training procedure consists of 45,000 iterations. 
We use a batch size of 256, AdamW~\citep{adamw} optimizer with a learning rate of $1e-3$. 
To compute the spectrogram consistent with both the decoder dimensions and the patch length, the number of coefficient bins is set to 255 and the window length to 40.
The masking ratio is set to 0.4. 
The masking ratio and the length of the training process have been determined by a sensitivity study (See Section \ref{sec:abla}).
The training procedure and the evaluations are performed on a desktop computer, with a Nvidia GeForce RTX 3070 GPU.

\section{Evaluation}\label{sec:eval}
To assess our hypotheses, we have conducted an extensive evaluation. 
This included five different baselines from key studies in single-lead ECG processing, three commonly used benchmarking datasets, and both linear probing and fine-tuning experiments. 
This section presents the details of the evaluation, along with the results and their discussion. 

\subsection{Baselines.} 

\begin{table*}[b]
\centering
\caption{Performance comparison of various methods under linear probing and fine-tuning on three ECG datasets. 
Best values in \textbf{bold} and second best \underline{underlined}.}
\label{tab:eval}
\resizebox{\textwidth}{!}{%
\begin{tabular}{lccccccccc}
\toprule
& \multicolumn{3}{c}{\textbf{MIT-AFIB}} 
& \multicolumn{3}{c}{\textbf{LT-AF}} 
& \multicolumn{3}{c}{\textbf{Physionet 2017}} \\ 
\cmidrule(lr){2-4}\cmidrule(lr){5-7}\cmidrule(lr){8-10}
\textbf{Methods} 
& \textbf{Accuracy} 
& \textbf{F1} 
& \textbf{AUC} 
& \textbf{Accuracy} 
& \textbf{F1} 
& \textbf{AUC} 
& \textbf{Accuracy} 
& \textbf{F1} 
& \textbf{AUC} \\
\midrule
\multicolumn{10}{c}{\textbf{Linear Probing}} \\
\midrule
\textbf{PCLR} 
& 0.691 $\pm$ 0.113 
& 0.636 $\pm$ 0.095 
& 0.731 $\pm$ 0.113 
& 0.808 $\pm$ 0.058 
& 0.568 $\pm$ 0.048 
& 0.869 $\pm$ 0.034 
& 0.6454 $\pm$ 0.0104
& 0.5597 $\pm$ 0.0094
& 0.7621 $\pm$ 0.0129 \\

\textbf{Mix-Up} 
& 0.691 $\pm$ 0.056 
& 0.579 $\pm$ 0.232 
& 0.746 $\pm$ 0.103 
& 0.813 $\pm$ 0.045 
& 0.593 $\pm$ 0.060 
& 0.889 $\pm$ 0.031
& 0.6546 $\pm$ 0.0186
& 0.5817 $\pm$ 0.0188
& 0.7906 $\pm$ 0.0109 \\

\textbf{DEAPS} 
& 0.790 $\pm$ 0.079 
& 0.687 $\pm$ 0.189 
& 0.851 $\pm$ 0.100 
& 0.829 $\pm$ 0.045 
& 0.592 $\pm$ 0.051 
& 0.893 $\pm$ 0.033
& 0.6791 $\pm$ 0.0126
& 0.6228 $\pm$ 0.0081
& 0.7897 $\pm$ 0.0082 \\

\textbf{CLOCS} 
& 0.680 $\pm$ 0.088 
& 0.567 $\pm$ 0.218 
& 0.709 $\pm$ 0.108 
& 0.748 $\pm$ 0.023 
& 0.534 $\pm$ 0.032 
& 0.843 $\pm$ 0.032 
& 0.6117 $\pm$ 0.0112
& 0.4650 $\pm$ 0.0186
& 0.7473 $\pm$ 0.0135 \\

\textbf{MTAE} 
& \underline{0.808 $\pm$ 0.937} 
& \underline{0.758 $\pm$ 0.071} 
& \underline{0.878 $\pm$ 0.080} 
& \underline{0.875 $\pm$ 0.035} 
& \underline{0.636 $\pm$ 0.043} 
& \underline{0.919 $\pm$ 0.024} 
& 0.6064 $\pm$ 0.0117
& 0.4820 $\pm$ 0.0056
& 0.7623 $\pm$ 0.0106 \\

\textbf{CuPID (Ours)} 
& \textbf{0.860 $\pm$ 0.041} 
& \textbf{0.782 $\pm$ 0.107} 
& \textbf{0.933 $\pm$ 0.010} 
& \textbf{0.880 $\pm$ 0.030} 
& \textbf{0.671 $\pm$ 0.042} 
& \textbf{0.934 $\pm$ 0.021}
& \textbf{0.7119 $\pm$ 0.0102}
& \textbf{0.6611 $\pm$ 0.0150}
& \textbf{0.8108 $\pm$ 0.0064} \\
\midrule
\multicolumn{10}{c}{\textbf{Fine-Tuning}} \\
\midrule
\textbf{PCLR} 
& 0.779 $\pm$ 0.097 
& 0.728 $\pm$ 0.098 
& \underline{0.897 $\pm$ 0.031} 
& 0.867 $\pm$ 0.035 
& 0.581 $\pm$ 0.025 
& 0.840 $\pm$ 0.036 
& 0.760 $\pm$ 0.012 
& 0.714 $\pm$ 0.012 
& 0.839 $\pm$ 0.012 \\

\textbf{Mix-Up} 
& 0.726 $\pm$ 0.067 
& 0.656 $\pm$ 0.052 
& 0.848 $\pm$ 0.071 
& 0.849 $\pm$ 0.042 
& 0.567 $\pm$ 0.031 
& 0.801 $\pm$ 0.034 
& 0.809 $\pm$ 0.011 
& 0.777 $\pm$ 0.013 
& 0.872 $\pm$ 0.002 \\

\textbf{DEAPS} 
& \underline{0.778 $\pm$ 0.079} 
& \underline{0.717 $\pm$ 0.074} 
& 0.873 $\pm$ 0.041 
& \underline{0.887 $\pm$ 0.031} 
& \textbf{0.626 $\pm$ 0.058} 
& \textbf{0.898 $\pm$ 0.039} 
& \textbf{0.822 $\pm$ 0.011} 
& \textbf{0.800 $\pm$ 0.019} 
& \textbf{0.871 $\pm$ 0.011} \\

\textbf{CLOCS} 
& 0.721 $\pm$ 0.072 
& 0.657 $\pm$ 0.092 
& 0.708 $\pm$ 0.126 
& 0.850 $\pm$ 0.039 
& 0.612 $\pm$ 0.067 
& 0.858 $\pm$ 0.043 
& 0.735 $\pm$ 0.013 
& 0.676 $\pm$ 0.021 
& 0.826 $\pm$ 0.013 \\

\textbf{MTAE} 
& 0.757 $\pm$ 0.077 
& 0.714 $\pm$ 0.035 
& 0.816 $\pm$ 0.047 
& 0.879 $\pm$ 0.031 
& 0.611 $\pm$ 0.023 
& 0.852 $\pm$ 0.037 
& 0.674 $\pm$ 0.018 
& 0.585 $\pm$ 0.034 
& 0.789 $\pm$ 0.011 \\

\textbf{CuPID (Ours)} 
& \textbf{0.888 $\pm$ 0.074} 
& \textbf{0.860 $\pm$ 0.081} 
& \textbf{0.955 $\pm$ 0.017} 
& \textbf{0.889 $\pm$ 0.024} 
& \underline{0.616 $\pm$ 0.040} 
& \textbf{0.887 $\pm$ 0.044} 
& \underline{0.805 $\pm$ 0.017} 
& \underline{0.773 $\pm$ 0.023} 
& \textbf{0.876 $\pm$ 0.012} \\
\bottomrule

\end{tabular}%
}
\end{table*}

%To assess the ability of the method to generalize different classes within the same record, given a limited number of labeled noisy recordings from Holter monitors, 

\ac{CuPID} has been evaluated against the following methods that compel the set of baselines for the evaluation; 
\ac{CLOCS} \citep{clocs},
\ac{PCLR} \citep{pclr},
\ac{MTAE}, from \cite{maefe};
\ac{DEAPS} \citep{deaps}; and
Mix-up \citep{mix_up}.
To ensure fairness in the evaluation, all the methods have been trained using the same optimizer for the same number of epochs.
The hyperparameter configuration for each baseline has been set up according to the specifics of each paper.
%
% To ensure fairness in the evaluation, all the methods have been trained using the same training configuration, encoder, and dataset as \ac{CuPID}. 
%

\subsection{Pre-Training Dataset}
The different methods necessitate specific properties in the pre-training dataset. 
For instance, PCLR requires ECG recordings from the same patient across different years, while DEAPS needs recordings longer than 2 minutes. 
Since all methods should be pre-trained using the same dataset to ensure a fair evaluation, 
the SHHS dataset \ac{SHHS} \citep{shhs1, shhs2} is used as the only pre-training data source. 
It has been identified as the only large publicly available dataset that meets the conditions for all methods used in the evaluation.
The details of \ac{SHHS} are provided in the Appendix \ref{sec:app_shhs}.
\\

\subsection{Downstream Datasets.} 
This section describes the datasets used for the various downstream tasks. These datasets, commonly employed for benchmarking ECG methods, have been carefully selected to offer different perspectives on the different method's performance, as detailed below.
All these databases are publicly available on Physionet \citep{physionet}. 

\paragraph{\acf{MIT-AFIB} \citep{mit-afib}:}
This dataset accommodates long-term recordings of 23 subjects transitioning between \ac{NSR} to paroxysmal \ac{AFib} episodes and vice versa.
To achieve strong performance on this downstream task, the encoder is expected to learn detailed representations during the pre-training step to discretize between the different states the subject experiences throughout the recording. This is necessary because the limited number of subjects seems insufficient for the encoder to learn these distinctions during the fine-tuning procedure.
%
%We want to highlight that \ac{CuPID} outperforms significantly all the baselines, as reflected in Table \ref{tab:eval1}.
%

\paragraph{\acf{LT-AF} \citep{l-af-dataset}:}
This dataset compels long-term recordings of 84 subjects. 
It is composed of subjects suffering spontaneous bradycardia episodes and subjects with sustained \ac{AFib} in addition to subjects suffering paroxysmal \ac{AFib} episodes that are also contained in the previous dataset.
This dataset includes more subjects than the previous one, although the number of subjects remains limited. 
It provides a different view of the method's performance since it introduces a new class of arrhythmia and is unbalanced. 
All this underscores the importance of pre-training. The encoder, similar to the previous dataset, cannot be expected to learn new features during fine-tuning that distinguish between the three classes.

%
%Table \ref{tab:eval1} reflects that \ac{CuPID} remarkably outperforms all the baselines. 

\noindent \textbf{Physionet Challenge 2017 \citep{physio_challenge}:}
This dataset comprises over 8,000 recordings aimed at distinguishing between Sinus Rhythm, AF, and other arrhythmias. 
Despite having the fewest instances, it can be assumed that each instance corresponds to a different subject. Existing literature \cite{ecg_size} indicates that ECG datasets scale more with the number of subjects rather than the number of instances per subject. 
Therefore, this downstream task will provide additional insights into how the performance of different methods scales when fine-tuning with a sufficiently large dataset.

\begin{table*}[!b]
\centering
\caption{Ablation study results for different parameter settings on three ECG datasets. The final row represents the mean MAE and CuPID across all experiments. Accuracy is the metric chosen for displaying the results.}
\label{tab:ablation-study}
\resizebox{\textwidth}{!}{%
\begin{tabular}{cccccccc}
\toprule
\multicolumn{2}{c}{\textbf{Hyperparameters}} & \multicolumn{2}{c}{\textbf{MIT-AFIB}} & \multicolumn{2}{c}{\textbf{LF-AF}} & \multicolumn{2}{c}{\textbf{Physionet 2017}} \\ 
\cmidrule(lr){1-2} \cmidrule(lr){3-4} \cmidrule(lr){5-6} \cmidrule(lr){7-8}
\textbf{PS} & \textbf{MR} & \textbf{MTAE} & \textbf{CuPID} & \textbf{MTAE} & \textbf{CuPID} & \textbf{MTAE} & \textbf{CuPID} \\ 
\midrule
10  & 0.3 & 0.733 $\pm$ 0.050 & \underline{0.776 $\pm$ 0.093} & 0.830 $\pm$ 0.040 & \underline{0.875 $\pm$ 0.034} & 0.648 $\pm$ 0.017 & \underline{0.684 $\pm$ 0.017} \\
10  & 0.4 & 0.777 $\pm$ 0.068 & \underline{0.793 $\pm$ 0.094} & 0.819 $\pm$ 0.035 & \underline{0.876 $\pm$ 0.030} & \underline{0.694 $\pm$ 0.014} & 0.681 $\pm$ 0.014 \\
10  & 0.5 & 0.656 $\pm$ 0.064 & \underline{0.821 $\pm$ 0.075} & 0.817 $\pm$ 0.041 & \underline{0.856 $\pm$ 0.034} & 0.658 $\pm$ 0.025 & \underline{0.676 $\pm$ 0.025} \\
10  & 0.6 & \underline{0.788 $\pm$ 0.070} & 0.751 $\pm$ 0.101 & 0.848 $\pm$ 0.028 & \underline{0.855 $\pm$ 0.049} & 0.615 $\pm$ 0.018 & \underline{0.686 $\pm$ 0.018} \\
10  & 0.7 & \underline{0.803 $\pm$ 0.076} & 0.737 $\pm$ 0.102 & \underline{0.835 $\pm$ 0.043} & 0.808 $\pm$ 0.064 & \underline{0.653 $\pm$ 0.012} & 0.636 $\pm$ 0.012 \\ 
\midrule
20  & 0.3 & 0.703 $\pm$ 0.054 & \underline{0.780 $\pm$ 0.078} & 0.852 $\pm$ 0.039 & \underline{0.889 $\pm$ 0.029} & \underline{0.685 $\pm$ 0.013} & 0.683 $\pm$ 0.013 \\
20  & 0.4 & 0.808 $\pm$ 0.093 & \underline{0.860 $\pm$ 0.041} & 0.875 $\pm$ 0.035 & \underline{0.880 $\pm$ 0.030} & 0.703 $\pm$ 0.012 & \underline{0.715 $\pm$ 0.012} \\
20  & 0.5 & 0.815 $\pm$ 0.070 & \underline{0.836 $\pm$ 0.090} & 0.870 $\pm$ 0.040 & \underline{0.882 $\pm$ 0.026} & 0.701 $\pm$ 0.012 & \underline{0.727 $\pm$ 0.012} \\
20  & 0.6 & \underline{0.819 $\pm$ 0.087} & 0.812 $\pm$ 0.086 & 0.860 $\pm$ 0.324 & \underline{0.876 $\pm$ 0.038} & 0.678 $\pm$ 0.011 & \underline{0.709 $\pm$ 0.011} \\
20  & 0.7 & 0.723 $\pm$ 0.093 & \underline{0.752 $\pm$ 0.072} & \underline{0.857 $\pm$ 0.025} & 0.849 $\pm$ 0.024 & 0.683 $\pm$ 0.024 & \underline{0.695 $\pm$ 0.024} \\ 
\midrule
25  & 0.3 & 0.768 $\pm$ 0.068 & \underline{0.833 $\pm$ 0.080} & 0.835 $\pm$ 0.043 & \underline{0.868 $\pm$ 0.051} & 0.670 $\pm$ 0.015 & \underline{0.697 $\pm$ 0.015} \\
25  & 0.4 & 0.793 $\pm$ 0.091 & \underline{0.821 $\pm$ 0.057} & 0.846 $\pm$ 0.041 & \underline{0.852 $\pm$ 0.043} & 0.673 $\pm$ 0.013 & \underline{0.693 $\pm$ 0.013} \\
25  & 0.5 & \underline{0.818 $\pm$ 0.050} & 0.815 $\pm$ 0.076 & 0.854 $\pm$ 0.040 & \underline{0.869 $\pm$ 0.045} & \underline{0.732 $\pm$ 0.013} & 0.727 $\pm$ 0.013 \\
25  & 0.6 & 0.799 $\pm$ 0.066 & \underline{0.860 $\pm$ 0.050} & 0.861 $\pm$ 0.044 & \underline{0.870 $\pm$ 0.036} & \underline{0.714 $\pm$ 0.024} & 0.709 $\pm$ 0.024 \\
25  & 0.7 & 0.773 $\pm$ 0.088 & \underline{0.810 $\pm$ 0.102} & 0.847 $\pm$ 0.010 & \underline{0.863 $\pm$ 0.040} & 0.667 $\pm$ 0.014 & \underline{0.695 $\pm$ 0.014} \\
\midrule
\multicolumn{2}{c}{\textbf{Mean ($\Delta$)}} & 0.772 $\pm$ 0.073 & 0.804 $\pm$ 0.080 ($\uparrow$4.2\%) & 0.847 $\pm$ 0.055 & 0.865 $\pm$ 0.038 ($\uparrow$2.1\%) & 0.679 $\pm$ 0.016 & 0.694 $\pm$ 0.016 ($\uparrow$2.3\%) \\

\bottomrule
\end{tabular}%
}
\end{table*}

\subsection{Experiments}
We carried out a 5-fold cross-validation for \ac{MIT-AFIB} and \ac{LT-AF} datasets. The training, validation, and test datasets were split with a ratio of 60-20-20, ensuring no patient overlap between the different partitions. It will artificially boost the performance. For the Physionet Challenge 2017 dataset, we adhered to the recommended train-test split.

In practical applications, just a single encoder should be used to derive useful representations that can be utilized across multiple downstream tasks to meet the real time monitoring efficiency needs. Consequently, we argue that linear probing evaluation holds greater importance, as fine-tuning the model creates task-specific encoders, requiring a distinct model for each task. 

However, to scientifically assess CuPID, we have conducted both linear probing and fine-tuning experiments for all the previously mentioned datasets.
For linear probing evaluation, a Logistic Regression model has been fitted on top of the representations. For the fine-tuning evaluation, the encoder weights were updated using an Adam \cite{adam} optimizer with a learning rate of 1e-4. The fine-tuning training finishes with an early-stopping patience of 5 based on the loss function values on the validation split, ensuring no decisions were influenced by the test split performance.

\subsection{Discussion of the Results}
Table \ref{tab:eval} illustrates that CuPID excels in all metrics across every downstream task in the linear proving evaluation. These results are noteworthy because, as previously mentioned, effective remote monitoring requires a single encoder capable of handling a diverse range of downstream tasks.

Regarding CuPID fine-tuning results, Table \ref{tab:eval} indicates that the performance improvement for datasets \ac{MIT-AFIB} and \ac{LT-AF} is minimal compared to linear probing. These findings align with recent studies \cite{finetuning}, which suggest that in highly limited cohort conditions, fine-tuning can yield worse results than linear probing. As previously mentioned, these outcomes are supported by research \cite{ecg_size}, which demonstrates that an ECG dataset scales with the number of patients rather than the number of instances. Notably, these datasets consist of 23 and 84 patients, respectively. Conversely, CuPID's performance improves when the dataset is composed of a larger number of subjects, such as Physionet Challenge 2017. CuPID outperforms the baseline set on most metrics, and delivers competitive results for the rest of them.

Lastly, we would like to highlight that CuPID consistently outperforms its analogue version, MTAE, which does not leverage the decoder with the contextual information provided by the spectrogram. A more detailed comparison between these two methods will be provided in the next section.  
These findings provide robust evidence in favor of the hypotheses posited by this study:
(I) The decoder’s inability to reconstruct the original signal due to the irregularity in the heartbeat intervals limits the encoder’s learning potential.
(ii) Cueing the decoder with the spectrogram endows it with the potential to solve the task. It drives the encoder to compute detailed patch representations which can be used to accurately reconstruct the input.
(iii) The more informative the patch representations are, the more performant the encode will be at the time of addressing downstream tasks.

\section{Ablation and Sensitivity Studies}\label{sec:abla}
This section aims to evaluate the impact of various hyperparameters on model performance and to determine whether incorporating the spectrogram in the decoder offers benefits across different hyperparameter configurations. Specifically, we will assess the influence of patch size, masking ratio, training duration, and encoder size. For all evaluated hyperparameter combinations, CuPID's performance will be compared to its analogue version without the incorporation of the spectrogram, MTAE. This evaluation will be conducted across the three downstream tasks. Due to the impracticality of fine-tuning numerous configurations, this evaluation will be conducted using linear probing. However recent literature \cite{cookbook} suggests is a valuable indicator of the quality of the representations.

\paragraph{Influence of the Masking Ratio and Patch Size:} Table \ref{tab:ablation-study} presents the performance of CuPID and MTAE across various configurations for the downstream datasets. These results validate the selection of a patch size of 20 and a masking ratio of 0.4 for both methods, as this combination consistently yields the best overall metrics. Notably, CuPID outperforms MTAE in 34 out of 45 possible combinations (over 75\%). This results in overall improvements ranging from 2.1\% to 4.2\% for each dataset. Additionally, CuPID achieves the best metrics for each dataset.

\paragraph{Influence of the Training Procedure Duration: }
We have evaluated the impact of pre-training duration. Figure \ref{fig:metrics_across_iterations} illustrates that performance in various downstream tasks improves up to the 45K iteration, after which a general decline is observed across different datasets for both methods. These findings support the selection of this iteration count and align with relevant studies \cite{mae} indicating that model performance deteriorates beyond a certain point. Notably, as shown in Figure \ref{fig:metrics_across_iterations}, CuPID outperforms in most checkpoints evaluated during the pre-training process across different datasets.

\begin{figure}[h]
\centering
\fbox{\includegraphics[width=\linewidth]{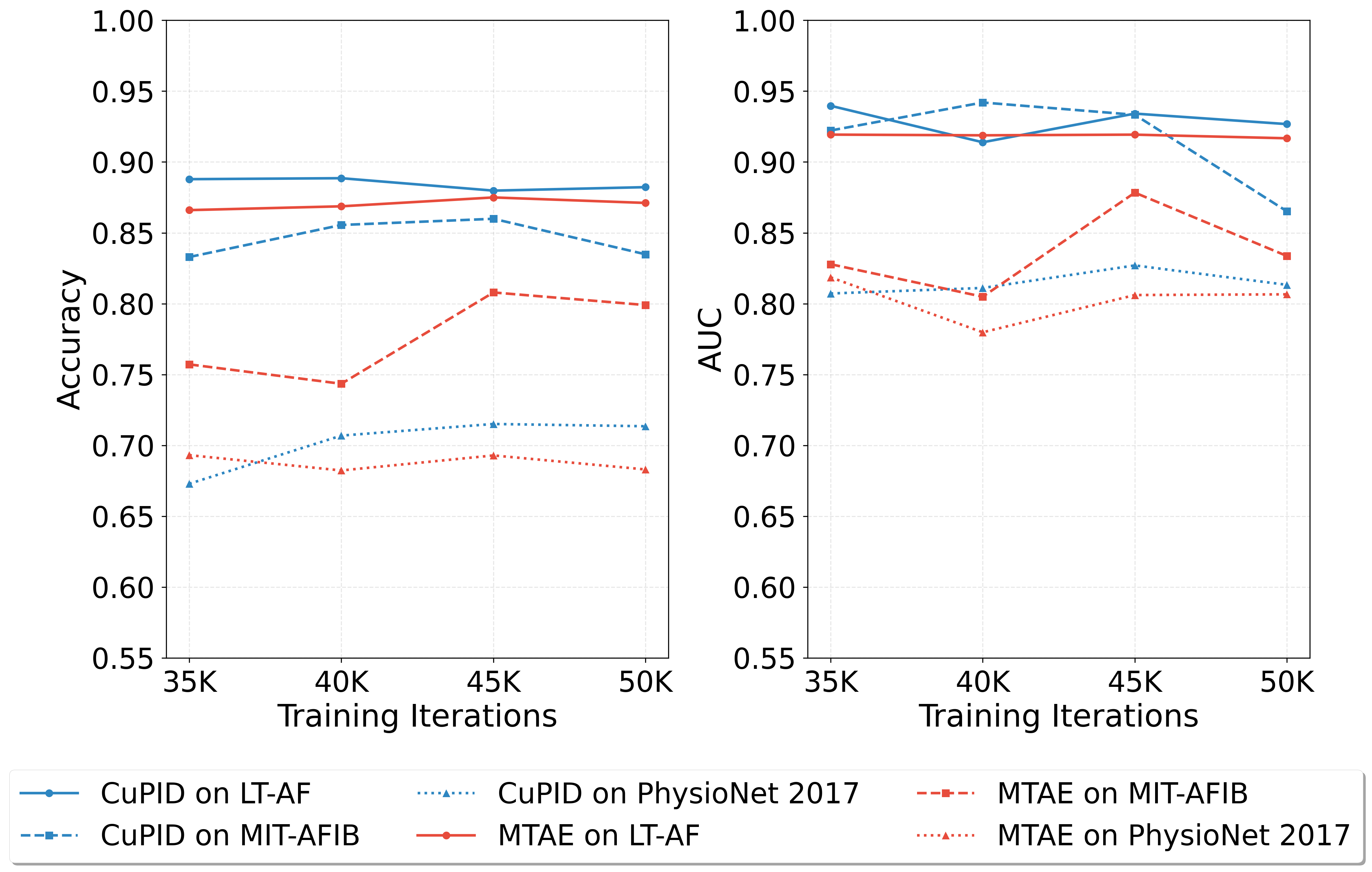}}
\caption{Evolution of the performance over training procedure}
\label{fig:metrics_across_iterations}
\end{figure}

\paragraph{Influence of the Model Size: }
We evaluated the performance of both models with various encoder sizes. Table 5 indicates that models larger than the proposed one do not show significant improvement. This suggests that an encoder with 4 transformer blocks is adequate to capture the complexity of single-lead ECG signals. Notably, CuPID consistently achieves better results across different datasets and model sizes.

\begin{figure}[ht]
\centering
\fbox{\includegraphics[width=0.9\linewidth]{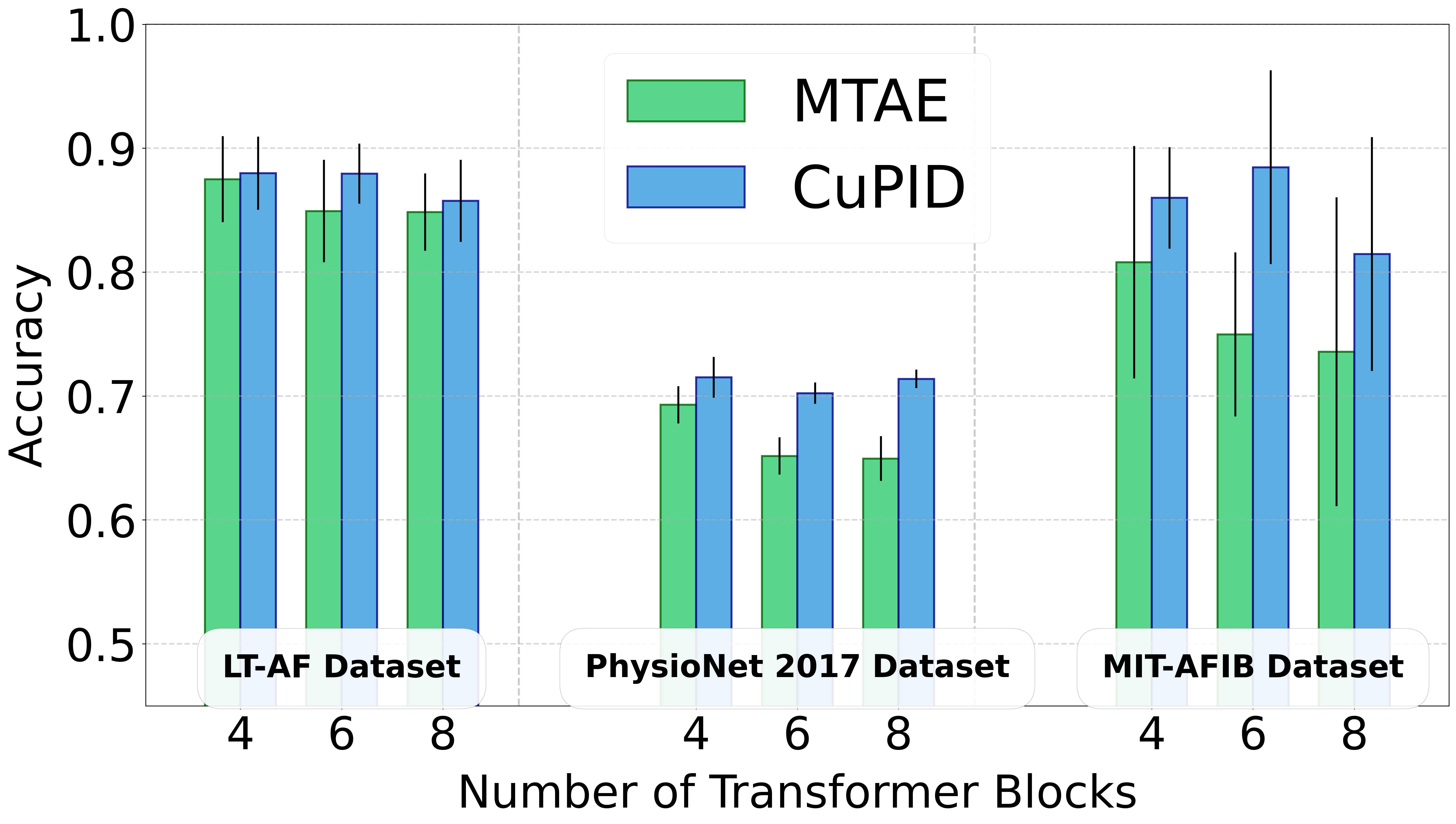}}
\caption{Impact of the model size on the model.}
\label{fig:metrics_across_iterations}
\end{figure}

\newpage
\section{Conclusion}
This research provides strong evidence that directly applying the \acf{MDM} framework to single-lead \ac{ECG} signals is suboptimal. 
Due to the lack of contextual information provided by the unmasked patches of the other leads, the baseline decoder faces challenges when handling irregular heartbeat intervals.
This leads the predictor to be cautious when reconstructing the masked patches and to not drive the encoder to compute detailed patch representations that can be used for addressing downstream tasks.
To overcome this issue, we introduce \ac{CuPID}, a novel \ac{SSL} technique for ECG analysis. By cueing the predictor with the contextual information given by the spectrogram of the input signal, \ac{CuPID} enforces the encoder to compute more informative representations. It results in a significant performance improvement when addressing downstream tasks. 

\paragraph{Limitations: } We have evaluated \ac{CuPID} solely using one pre-training dataset. As discussed in the manuscript, this is due to \ac{SHHS} is the only dataset that fits the requirements of all the \ac{SSL} methods that compose set of baselines.

\section{Reproducibility Statement}
The attached code as a part of the supplementary material encompasses the implementation of \ac{CuPID} and several other baselines. Moreover, comprehensive details on training hyperparameters, schemes, and hardware specifications are provided. In addition the pseudocode for the method is provided in the Appendix. Finally, we furnish the pre-trained model's parameters to facilitate others in achieving reproducible results, together with the code used for processing each database.

\newpage

\newpage

\bibliography{example_paper}
\bibliographystyle{icml2025}

%%%%%%%%%%%%%%%%%%%%%%%%%%%%%%%%%%%%%%%%%%%%%%%%%%%%%%%%%%%%%%%%%%%%%%%%%%%%%%%
%%%%%%%%%%%%%%%%%%%%%%%%%%%%%%%%%%%%%%%%%%%%%%%%%%%%%%%%%%%%%%%%%%%%%%%%%%%%%%%
% APPENDIX
%%%%%%%%%%%%%%%%%%%%%%%%%%%%%%%%%%%%%%%%%%%%%%%%%%%%%%%%%%%%%%%%%%%%%%%%%%%%%%%
%%%%%%%%%%%%%%%%%%%%%%%%%%%%%%%%%%%%%%%%%%%%%%%%%%%%%%%%%%%%%%%%%%%%%%%%%%%%%%%
\newpage
\appendix
\label{sec:appendix}
\onecolumn
\section{Data Preprocessing} \label{app:data}
To ensure the complete reproducibility of this work, this section presents a detailed description of the preprocessing steps employed for the training and evaluation databases utilized in the proposed method.

\subsection{\ac{SHHS} Data Selection}
Only the subjects that appear in both recording cycles are used during the training procedure. This leads to 2643 subjects. ECG signals are extracted from the \ac{PSG} recordings. The quality of every 10 seconds-data strips has been evaluated with the algorithm proposed by Zhao and Zhang \cite{ecg_quality}. We use SHHS since it contains two records belonging to the same subject. This makes this specific database special, and this is the reason that it has been the only database used during the optimization.

\subsection{Data cleaning}
In addition, all signals from the utilized datasets were resampled to a frequency of 100Hz. Then, a $5^{th}$ order Butterworth high-pass filter with a cutoff frequency of 0.5Hz was applied to eliminate any DC-offset and baseline wander. Finally, each dataset underwent normalization to achieve unit variance, ensuring that the signal samples belong to a $\mathcal{N}(0, 1)$ distribution. This normalization process aimed to mitigate variations in device amplifications.

\section{Details of SHHS Dataset} 
\label{sec:app_shhs}

\begin{table}[!h]
\centering
\caption{Demographic and Clinical Distribution in the SHHS Dataset}
\label{tab:shhs-stats}
\small % Reduce font size to save space
\renewcommand{\arraystretch}{0.8} % Reduce row height
\resizebox{0.5\textwidth}{!}{ % Reduce overall table width
\begin{tabular}{lc}
\toprule
\textbf{Category} & \textbf{Percentage (\%)} \\
\midrule
\textbf{Prevalent Atrial Fibrillation (AFib)} & \\
\hspace{3mm} No & 49.8 \\
\hspace{3mm} Yes & 0.8 \\
\hspace{3mm} Unknown & 49.4 \\
\midrule
\textbf{Race Distribution} & \\
\hspace{3mm} White & 84.5 \\
\hspace{3mm} Black & 8.9 \\
\hspace{3mm} Other & 6.6 \\
\midrule
\textbf{Age (at Visit 1)} & 63.1 $\pm$ 11.2 years \\
\midrule
\textbf{Sex Distribution} & \\
\hspace{3mm} Male & 47.6 \\
\hspace{3mm} Female & 52.4 \\
\bottomrule
\end{tabular}}
\end{table}

\section{Details of Datasets used for Main Evaluation of Single-Lead ECG Baselines } \label{sec:app-databases}

% \begin{table}[H]
% \centering
% \caption{\acf{MIT-AFIB}}
% \label{tab:mit-stats}
% \resizebox{\textwidth}{!}{%
% \begin{tabular}{lcccccc}
% \toprule

% Label   & \# ECGs & \# Record. Count \& (Ratio) & Ratio \#ECGs per Record. \\ \midrule 

% \acf{NSR}   &  50115   &  21 (91.3\%) &  0.401 $\pm$ 0.357 \\ \midrule
% \acf{AFib}  &  33694   &  23 (100\%)   & 0.656 $\pm$ 0.320  \\ \midrule 
% \end{tabular}}
% \end{table}

% \begin{table}[H]
% \centering
% \caption{\acf{LT-AF}}
% \label{tab:mit-stats}
% \resizebox{\textwidth}{!}{%
% \begin{tabular}{lcccccc}
% \toprule

% Label   & \# ECGs & \# Record. Count \& (Ratio) & Ratio \#ECGs per Record. \\ \midrule 

% \acf{NSR}   & 270702  &  53 (63.1\%) &  0.672 $\pm$ 0.315 \\ \midrule
% \acf{AFib}  & 368272 &  84 (100\%)   & 0.546 $\pm$ 0.422  \\ \midrule 
% Bradycardia  &  19197   &  35 (41.7\%)   & 0.072 $\pm$ 0.100  \\ \midrule 
% \end{tabular}}
% \end{table}

\begin{table}[!h]
\centering
\caption{Statistics of the PhysioNet 2017 Training Set}
\label{tab:physionet-2017}
\begin{tabular}{lccc}
\toprule
\textbf{Label} & \textbf{Total Recordings} & \textbf{Mean Time (s)} & \textbf{SD Time (s)} \\ 
\midrule
Normal        & 5,154  & 31.9  & 10.0  \\ 
AF            & 771    & 31.6  & 12.5  \\ 
Other rhythm  & 2,557  & 34.1  & 11.8  \\ 
Noisy         & 46     & 27.1  & 9.0   \\ 
\midrule
\textbf{Total} & \textbf{8,528} & \textbf{32.5} & \textbf{10.9} \\ 
\bottomrule
\end{tabular}
\end{table}

\begin{table}[!h]
\centering
\caption{Statistics of the \acf{MIT-AFIB} Database}
\label{tab:mit-stats}
\resizebox{0.8\textwidth}{!}{%
\begin{tabular}{lccc}
\toprule
\textbf{Label} & \textbf{Total ECGs} & \textbf{Record Count (\%)} & \textbf{Avg. ECGs per Record} \\ 
\midrule
\acf{NSR}  & 50,115  & 21 (91.3\%)  & 0.401 $\pm$ 0.357  \\ 
\acf{AFib} & 33,694  & 23 (100\%)   & 0.656 $\pm$ 0.320  \\ 
\bottomrule
\end{tabular}}
\end{table}

\begin{table}[!h]
\centering
\caption{Statistics of the \acf{LT-AF} Database}
\label{tab:lt-af-stats}
\resizebox{0.8\textwidth}{!}{%
\begin{tabular}{lccc}
\toprule
\textbf{Label} & \textbf{Total ECGs} & \textbf{Record Count (\%)} & \textbf{Avg. ECGs per Record} \\ 
\midrule
\acf{NSR}       & 270,702  & 53 (63.1\%)  & 0.672 $\pm$ 0.315  \\ 
\acf{AFib}      & 368,272  & 84 (100\%)   & 0.546 $\pm$ 0.422  \\ 
Bradycardia     & 19,197   & 35 (41.7\%)  & 0.072 $\pm$ 0.100  \\ 
\bottomrule
\end{tabular}}
\end{table}

\newpage
\subsection{Pseudocode}

\begin{algorithm}[!h]
   \caption{CuPID Training Algorithm}
   \label{alg:training}
\begin{algorithmic}
   \STATE {\bfseries Input:} Number of iterations $K$, Batch size $B$
   \STATE {\bfseries Input:} Encoder $\mathcal{F}(x)$, Predictor $\mathcal{P}(h, s)$
   \STATE {\bfseries Input:} Parameters $\theta$, Optimizer $\text{opt}(\theta, \nabla\theta)$
   \STATE {\bfseries Input:} Spectrogram Transform $S(x)$, Mask Function $\mathcal{RM}(X)$
   \STATE {\bfseries Input:} Mask Tokens $M_t$, Loss Function $\mathcal{L}_1(X, Y, M)$

   \FOR{$k=0$ {\bfseries to} $K$}
       \STATE Sample mini-batch: $X \gets \{X^1, \dots, X^N\}_{b=0}^{B}$
       \STATE Apply random masking: $(H_m, M) \gets \mathcal{RM}(X)$
       \STATE Compute encoder representations: $H_m \gets \mathcal{F}(H_m)$
       \STATE Attach mask tokens: $H \gets \text{Rec}(H_m, M_t)$
       \STATE Compute spectrogram: $S \gets S(X)$
       \STATE Compute predictor output: $Y \gets \mathcal{P}(H, S)$
       \STATE Compute loss: $l \gets \mathcal{L}_1(X, Y, M)$
       \STATE Compute gradients: $\nabla\theta \gets \frac{\partial l}{\partial \theta}$
       \STATE Update parameters: $\theta \gets \text{opt}(\theta, \nabla\theta)$
   \ENDFOR
\end{algorithmic}
\end{algorithm}

\begin{algorithm}[!h]
   \caption{CuPID Predictor Algorithm}
   \label{alg:predictor}
\begin{algorithmic}
   \STATE {\bfseries Input:} Predictor Layers $\mathcal{P}$, Final Layer $O(H)$
   \STATE {\bfseries Input:} Predictor Input $H$, Spectrogram $S$

   \FOR{each $(idx, P_t)$ in $\mathcal{P}$}
       \IF{$idx = 0$}
           \STATE $H \gets P_t(H, H, H)$
       \ELSE
           \STATE $H \gets P_t(H, S, H)$ \COMMENT{Feed Spectrogram as Key}
       \ENDIF
   \ENDFOR
   \STATE $Y \gets O(H)$
   \STATE \textbf{return} $Y$
\end{algorithmic}
\end{algorithm}

%--------------------------------------------------------------------------------------------------------------------------------------------

% \begin{algorithm*}
% \SetAlgoLined
% \KwIn{
%     \\ \hspace{0.5cm} $\mathcal{P}$, $\mathcal{O}(H)$ \Comment{Predictor and Final Layer}
%     \\ \hspace{0.5cm} H, S \Comment{Predictor Input and Spectrogram}
%     } 

% \vspace{0.25cm}
% \For{$idx, \mathcal{P}_{l}$  in $ enum(\mathcal{P})$} {  
      
%       \eIf{$idx = 0$}
%       {
%         $ H \gets \mathcal{P}_l(H, H, H)$\;
%       }
%       {
%         $ H \gets \mathcal{P}_l(H, S, H)$     \Comment{Fed the Sectrogram as the Key}   
%       }
%     }
%     $ Y \gets \mathcal{O}(H)$
%     return $Y$\;

% \caption{CuPID's Predictor}
% \label{algo}
% \end{algorithm*}

%%%%%%%%%%%%%%%%%%%%%%%%%%%%%%%%%%%%%%%%%%%%%%%%%%%%%%%%%%%%%%%%%%%%%%%%%%%%%%%
%%%%%%%%%%%%%%%%%%%%%%%%%%%%%%%%%%%%%%%%%%%%%%%%%%%%%%%%%%%%%%%%%%%%%%%%%%%%%%%

\end{document}